\documentclass[journal]{IEEEtran}
\usepackage{graphicx}
\usepackage{float}
\usepackage{subcaption}
\usepackage{amsmath}
\usepackage{bm}
\usepackage{upgreek}
\usepackage{color}
\usepackage{gensymb}
\usepackage{cite}
\usepackage[numbers,sort&compress]{natbib}
\usepackage[left=0.75in,right=0.75in,top=1in,bottom=0.8in]{geometry}
\usepackage{diagbox}
\usepackage{algorithm}  
\usepackage{algpseudocode}  
\usepackage{amsmath}  
\begin{document}

\title{Bio-inspired Neural Network-based Optimal Path Planning for UUVs under the Effect of \\Ocean Currents}

\author{Danjie Zhu,~\IEEEmembership{Student Member,~IEEE},~and~Simon X. Yang,~\IEEEmembership{Senior Member,~IEEE}
\thanks{This work was supported by the Natural Sciences and Engineering Research Council (NSERC) of Canada. \textit{(Corresponding author: Simon X. Yang.)} The authors are with Advanced Robotics and Intelligent System (ARIS) Laboratory, School of Engineering, University of Guelph, Guelph, ON. N1G2W1, Canada (e-mail: \{danjie; syang\}@uoguelph.ca).}
}

\maketitle

\begin{abstract}
To eliminate the effect of ocean currents when addressing the optimal path in the underwater environment, an intelligent algorithm designed for the unmanned underwater vehicle (UUV) is proposed in this paper. The algorithm consists of two parts: a neural network-based algorithm that deducts the shortest path and avoids all possible collisions; and an adjusting component that balances off the deviation brought by the effect of ocean currents. The optimization results of the proposed algorithm are presented in detail, and compared with the path planning algorithm that does not consider the effect of currents. Results of the comparison prove the effectiveness of the path planning method when encountering currents of different directions and velocities.
\end{abstract}

%
\IEEEpeerreviewmaketitle

\section{Introduction}
\IEEEPARstart
{A}{ttentions} have been attracted to the field of undersea exploration for decades due to the abundant resources embedded in the ocean area, such as the biological resources, mineral resources and space resources \cite{r1,r2,r4}. Among the technologies that are beneficial for underwater exploration, the unmanned underwater vehicle (UUV) detection is one of the most significant and applicable technologies. To complete the vehicle’s underwater operation, path planning problem of the UUV is worthy of thoroughly studying \cite{r5,RN1346}. Considering the complex condition of the deep-water area, such as the invisibility and irregular obstacles, the ability for the UUV to navigate with an in-time reaction to avoid collisions and address the most optimal path between the starting and ending point, is very important \cite{RN1338}. Generally, the path planning process of the UUV is to start from an initial position, then search the shortest path or in the shortest time to reach the destination \cite{reviewuuvpp}. During the searching process all possible obstacles should be avoided. In the underwater environment where ocean currents commonly exist, the UUV may deviate from the desired path due to the effect of currents, thus leading to the wastage of time and energy \cite{3dppauv,pptime}. For example, the vehicle may navigate along a farther path or consume longer time if the ocean currents come in the direction that drags the vehicle away from the destination. 

Various intelligent algorithms have been developed to resolve the optimal path planning problem of the unmanned vehicle in the marine environment for a long time \cite{3dpp,ppcon,pptime}. Most of the algorithms are conducted on a grid-based modeling of the environment. The grid-based path planning method was first introduced in the form of the Dijkstra algorithm, where it needs to search all the possible paths, and address the shortest path solution between the origin node and the destination node when the latter is determined \cite{Dijkstra}. Then A* algorithm is applied on the basis of computing the heuristic cost, which increases the efficiency of the path planning process by largely reducing the searching space \cite{Astar1}. However, few studies have thoroughly considered the effect of environmental disturbances for UUV in the marine area, while most of them limited the discussion on the wind effect on the unmanned aerial vehicle (UAV) \cite{uavcurrent2,wind1,wind2,wind4}. The typical marine disturbances, such as the obstacles and currents, are usually distributed irregularly in the environment \cite{RN1345}. This affects the practical application of classic grid-based path planning methods, as they always require a pre-defined model with all the conditions clarified and hardly changed \cite{RN1338}.

Some researchers have refined the A* algorithm by involving the effect of environmental factors such as moving obstacles and ocean currents \cite{RN1337,RN1338}. They provide a feasible method for the vehicle to maintain safety and efficiency when tracking the desired path, yet they only discuss the application of the unmanned surface vehicle (USV), which is not highly applicable to the case of UUV path planning. The limited speed and undersea working place of the UUV make the planning process demanding on the real-time feedback from the environment and give reactions accordingly \cite{rr3}. To systematically analyze the effect of ocean currents in underwater path planning, Mengxue and his colleagues have developed a flow field partition method to extract the key fluid influence of the ocean currents \cite{RN1344}. They partition the dynamic flow field into a static partition of uniform flow, which simplifies the computation of complex states of the currents. Evolutionary algorithm (EA) has been applied in the underwater path planning problem to tackle the real-time feedback of the environmental disturbances, as the method can refine the optimization with the changing influence simultaneously \cite{EA1,RN1346,ant2}. However, EA usually requires higher computation complexity and larger amounts of data inputs \cite{EA2}. The bio-inspired neural network algorithm has been raised in recent years for planning the optimal path in the marine system. Compared with A* and EA methods, it satisfies the requirements of both in-time feedback to the changing environment and acceptable computation complexity. The bio-inspired neural network algorithm continuously updates the state of neurons by transmitting the information through the network to give an instant reaction, and reduces the complexity by limiting the searching area in a certain range \cite{bioins,RN1348}.

In this paper an intelligent algorithm based on the bio-inspired neural network algorithm for UUV path planning in the currents-affected marine environment is proposed and analyzed. By combining the neural network theory with the optimization by the distance and driving direction, the shortest path without collisions can be addressed. Meanwhile the algorithm that balances off the deviation induced by the effect of currents is added. Simulation results are illustrated in detail and comparisons for algorithms with or without considering the existence of ocean currents are also presented.

 
\section{The Proposed Path Planning Algorithm Design}
In this section, the modeling of the underwater environment is first presented. Then the proposed neural network for 2D and 3D modeling environment are defined, with the optimal path addressed by its transfer function. After that, the current effect elimination process is described and combined with the pre-defined neural network to form the final current effect-eliminated algorithm.
\subsection{Modeling of the underwater environment}
In this paper, the environment for the UUV is first considered as a 2D grid map, consisting of square cells of the same size (Fig. 1). The cells are decomposed as small enough to represent only two conditions: fully occupied by obstacles or not. The color of the cell represents its occupied condition, where black cells represent the existence of the obstacles and the vehicle cannot pass. To address the exact position of the vehicle during the planning process, a 2D coordinate system is established for the map. The side length of each cell is defined as one unit of the system. \begin{figure}[H]
\begin{center}
        \includegraphics[scale=0.44]{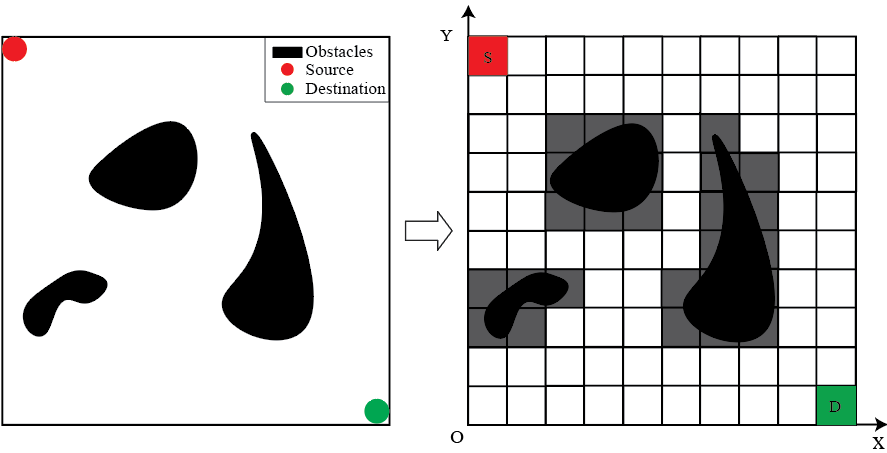}                         
        \caption*{Fig. 1. 2D grid map modeling in the underwater environment.}	
\end{center} 
\end{figure} Therefore each cell is assigned with a specific coordinate value for helping reduce the calculation complexity when building the neural network.

\subsection{Building the neural network}
In this section, the neural network models are established in both two dimensional and three dimensional conditions, based on the pre-defined grid-based coordinates. 
\subsubsection{2D Neural Network}
A 2D neural network is built on the 2D grid coordinate system modeled in the previous section (Fig. 2). Each cell is considered as a neuron of the neural network correspondingly, e.g. the $i^{th}$ cell represents the $i^{th}$ neuron. Each neuron can pass its information to the eight neighboring neurons, within the receptive range shown in Fig. 2. In other words, once a neuron is activated, its eight neighboring neurons are waited to be activated and receive the corresponding information. Then one of the neighboring neurons is activated and excites the next activation loop. This exactly corresponds to the path planning process of the UUV on the 2D grid map. When the vehicle reaches a cell, one of its eight neighboring cells will be the next location on the path. The choice of the next activated neuron as well as the next location cell is based on the intelligent algorithm raised in the following section.
\begin{figure}[ht]
\begin{center}
        \includegraphics[scale=0.235]{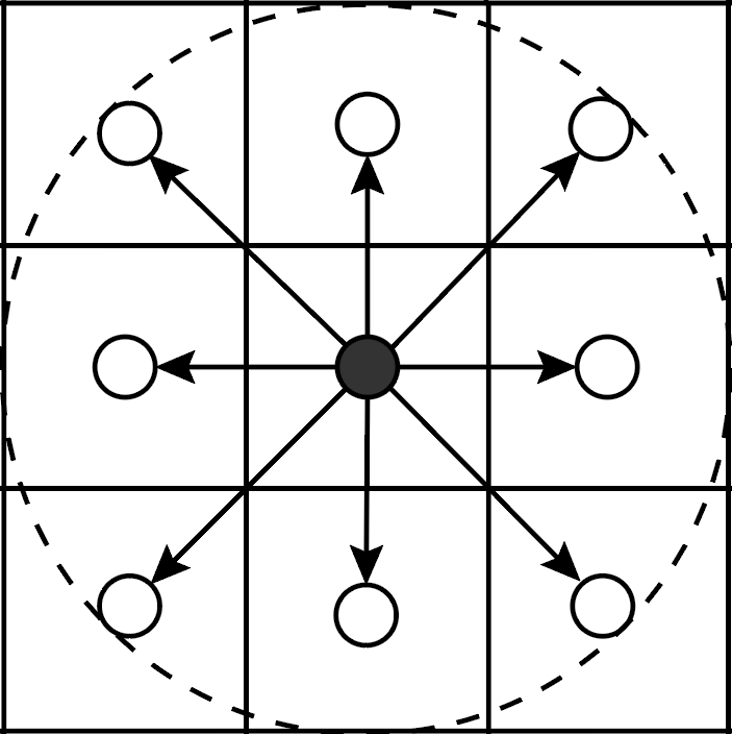}                         
        \caption*{Fig. 2. The 2D neural network.}	
\end{center} 
\end{figure}
\subsubsection{Extending to 3D Neural Network}
On the basis of the 2D grid-based neural network, a third dimension is introduced to establish the model of the path planning problem, which exactly corresponds to the practical case (see Fig. 3). Similarly, in the 3D neural network, each cube represents a neuron. The information each neuron carries can be passed to the 26 neighbouring neurons (in the white ball) that belong to the 26 cubes around it. The receptive range is determined by a sphere of $\sqrt{2}$ radius that centered at each neuron, covering the 26 neighbouring neurons (in the grey sphere).\begin{figure}[H]
\begin{center}
        \includegraphics[scale=0.345]{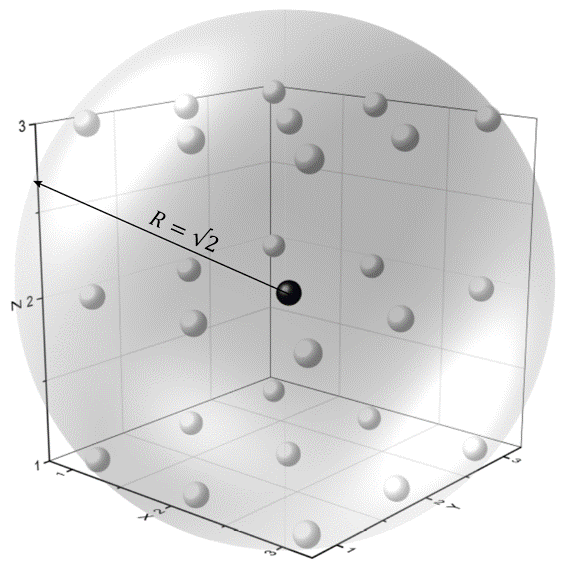}                         
        \caption*{Fig. 3. The 3D neural network (current neuron: (2,2,2)).}	
\end{center} 
\end{figure} One of the 26 neighbouring cubes will be the next location for the vehicle, which is chosen according to the neural network-based algorithm given in the paper.

\subsection{Current effect-eliminated Bio-inspired Neural Network Path Planning (CBNNP) Algorithm Design}
Established on the grid-based neural network model introduced in previous sections, the current effect-eliminated bio-inspired neural network path planning algorithm (CBNNP) is proposed. The CBNNP algorithm consists of two parts: (1) a bio-inspired neural network path planning (BNNP) algorithm component that deducts the optimal path by avoiding all the obstacles; (2) a correcting component that eliminates the deviation brought by the effect of ocean currents.
\subsubsection{Bio-inspired Neural Network Path Planning (BNNP) Algorithm}
The intelligent path planning algorithm can be derived based on the modeled 2D grid coordinates and the corresponding neural network. The operating mechanism of the algorithm is to deduct the optimal path that composed by the continuous coordinates of the vehicle movement. A discrete-time Hopfield-type neural network algorithm is applied to conduct the program to address the most optimal path \cite{Hoppp}. The starting and the ending positions are known, and the position and size of the obstacles are set randomly. Dynamic parameters such as the desired velocity value of the vehicle $\mathbf{v_d}$ are given. More importantly, the direction and the velocity of the currents $\mathbf{v_{cur}}$ are also known at the beginning. The dynamic model of the algorithm is described as
\begin{equation}
    a_i(t+1)=g(a_j(t)+e^{(-||i-D||)}+I_i)
\end{equation}

\noindent where $a_i (t+1)$ is the neural activity of the $i^{th}$ neuron at the time $t+1$; $a_j (t)$ is the neural activity of the $j^{th}$ neuron at the time $t$. 

Under the 2D condition, the number $i$ ranges from 1 to 8, which represents the eight neurons around the excited $j^{th}$ neuron within the receptive scope defined in the previous section (Fig. 2). Under the 3D condition, the number $i$ ranges from 1 to 26, which represents the twenty-six neurons around the excited $j^{th}$ neuron within the receptive scope defined in Fig. 3. Moreover, $||i-D||$ is the Euclidean distance between the $i^{th}$ neuron and the destination D. 
The transfer function $g(x)$ is
\begin{equation}
    g(x)=\left\{
    \begin{aligned}
    -1, \quad x<0\\
    k_gx, \quad x\geq0
    \end{aligned}
    \right.
\end{equation}
\noindent where $k_g$ is a constant that ranges from 0 to 1. 

The external input to the $i^{th}$ neuron $I_i$ is determined based on the obstacle occupied condition and the searched status of the current neuron as
\begin{equation}
    I_i=\left\{
    \begin{aligned}
    -1, \quad \text{if obstacles}\\
    0, \quad \text{if covered}\\
    +1, \quad \text{if uncovered}
    \end{aligned}
    \right.
\end{equation}

Under the 2D condition, each time when the vehicle is trying to address its next position, the 8 units around it will be considered as the next possible transferring neurons. Their coordinates can be represented by the addition or subtraction of 1 from the vehicle’s current coordinate. Under the 3D condition, the twenty-six cubes around the current neuron will be the next possible local solution.

\subsubsection{Elimination of the Current Effect}

The choice of the next position along the ideal path is based on the current position of the vehicle. In other words, the optimal path is formed step by step in the unit of a grid. Therefore, the effect of the ocean currents can be discretized in the unit of grids, as given in Fig. 4. It is clearly shown that the difference between the direction of the ocean currents velocity $\mathbf{v_{cur}}$ and the direction of the vehicle’s desired velocity $\mathbf{v_d}$ obviously affect the sailing trajectory of the vehicle at each position, thus resulting into the large deviation from the optimal path after the accumulation.\\ 
\indent To eliminate the deviation, a third party velocity vector given by the vehicle, the planned velocity vector $\mathbf{v_{plan}}$ is derived, based on the parallelogram law (see Fig. 4(b)). According to the parallelogram law, summation of $\mathbf{v_{cur}}$ and $\mathbf{v_{plan}}$ results into $\mathbf{v_d}$, which allows the vehicle to drive along the desired direction with the desired velocity. The currents velocity vector $\mathbf{v_{cur}}$ is obtained from the environment. Specific direction of the desired velocity vector $\mathbf{v_d}$ is derived by the optimal path, based on the intelligent path planning algorithm raised in this paper. Additionally, the size of $\mathbf{v_d}$ is set at the beginning. 

As the size of $\mathbf{v_d}$ and $\mathbf{v_{cur}}$ are known, according to the grid-based coordinate system, suppose the coordinates of $\mathbf{v_d}$ is ($x_d$, $y_d$), and the $\mathbf{v_{cur}}$ is ($x_{cur}$, $y_{cur}$), the position of the planned velocity vector $\mathbf{v_{plan}}$ can be addressed by these coordinates based on the parallelogram law as (see Fig. 4(b))
\begin{equation}
    \mathbf{v_{plan}}:(x_d-x_{cur}, y_d-y_{cur})
\end{equation}


\begin{figure}[ht]
\begin{center}
        \includegraphics[scale=0.34]{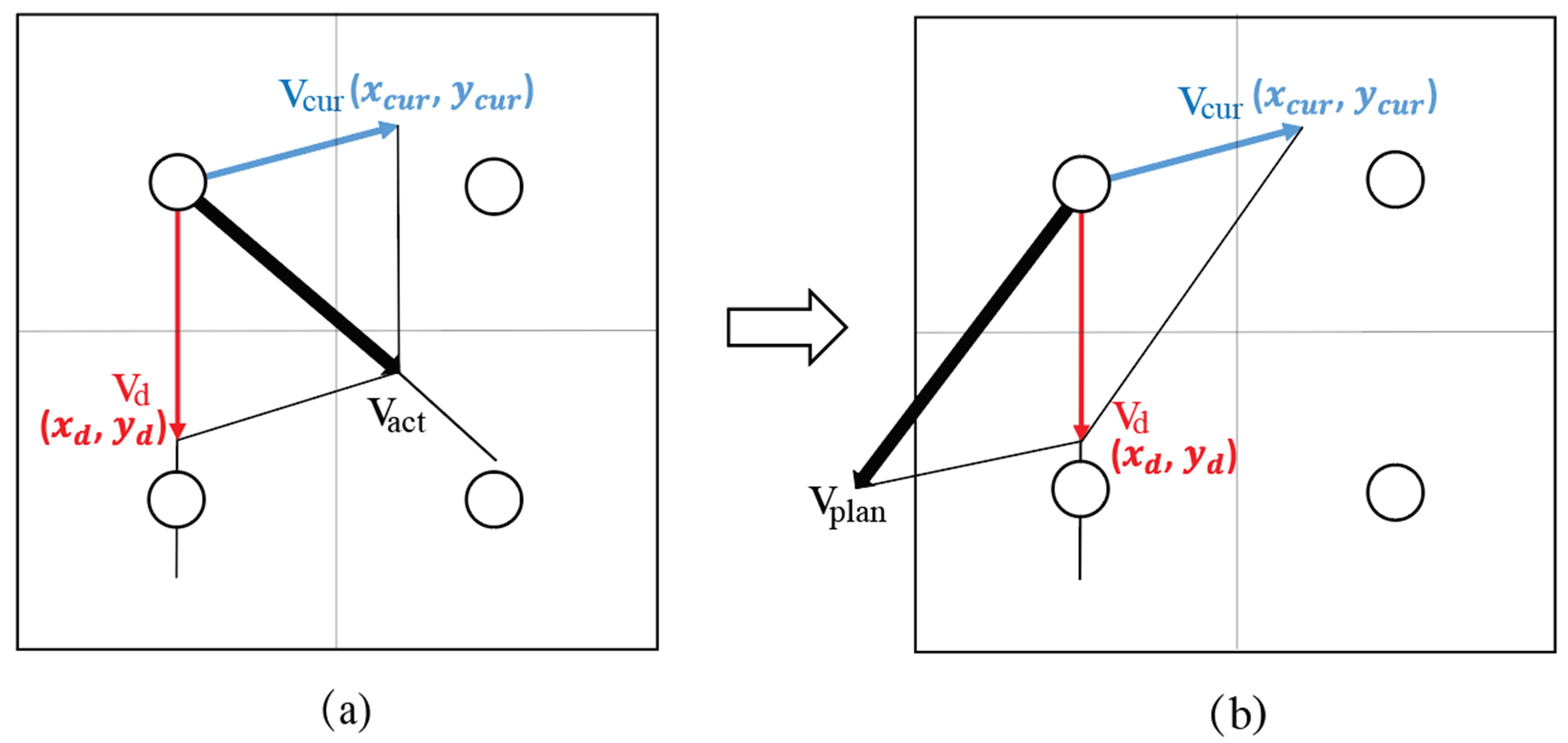}                         
        \caption*{Fig. 4. Elimination of the ocean current effect by introducing the third party vector $\mathbf{v_{plan}}$ in the 2D modeling. (a) deviation induced without $\mathbf{v_{plan}}$; (b) no deviation with $\mathbf{v_{plan}}$.}
\end{center} 
\end{figure}

When extending to the 3D model, similar mechanism is applied where the third party velocity vector $\mathbf{v_{plan}}$ is introduced , with an extra dimension added (see Fig. 5(b)). According to the parallelogram law, $\mathbf{v_{plan}}$ balances off the effect brought by the ocean currents vector $\mathbf{v_{cur}}$ and finally gives the summation result of the desired velocity vector $\mathbf{v_d}$ to follow the optimal planning path. Precise directions and size of $\mathbf{v_{plan}}$ are deducted by the 3D coordinates of the separate vectors. As the direction and speed of $\mathbf{v_d}$ and $\mathbf{v_{cur}}$ are given, suppose the 3D coordinates of $\mathbf{v_d}$: ($x_d$, $y_d$, $z_d$), $\mathbf{v_{cur}}$: ($x_{cur}$, $y_{cur}$, $z_{cur}$). The 3D coordinates of $\mathbf{v_{plan}}$ can be obtained on basis of the parallelogram relationship illustrated in the Fig. 5(b) as
\begin{equation}
    \mathbf{v_{plan}}: (x_d-x_{cur}, y_d-y_{cur}, z_d-z_{cur}).
\end{equation}
\indent Therefore $\mathbf{v_{plan}}$ is derived to complete the elimination of the ocean current effect in the 3D environment.

\begin{figure}[ht]
\begin{center}
        \includegraphics[scale=0.42]{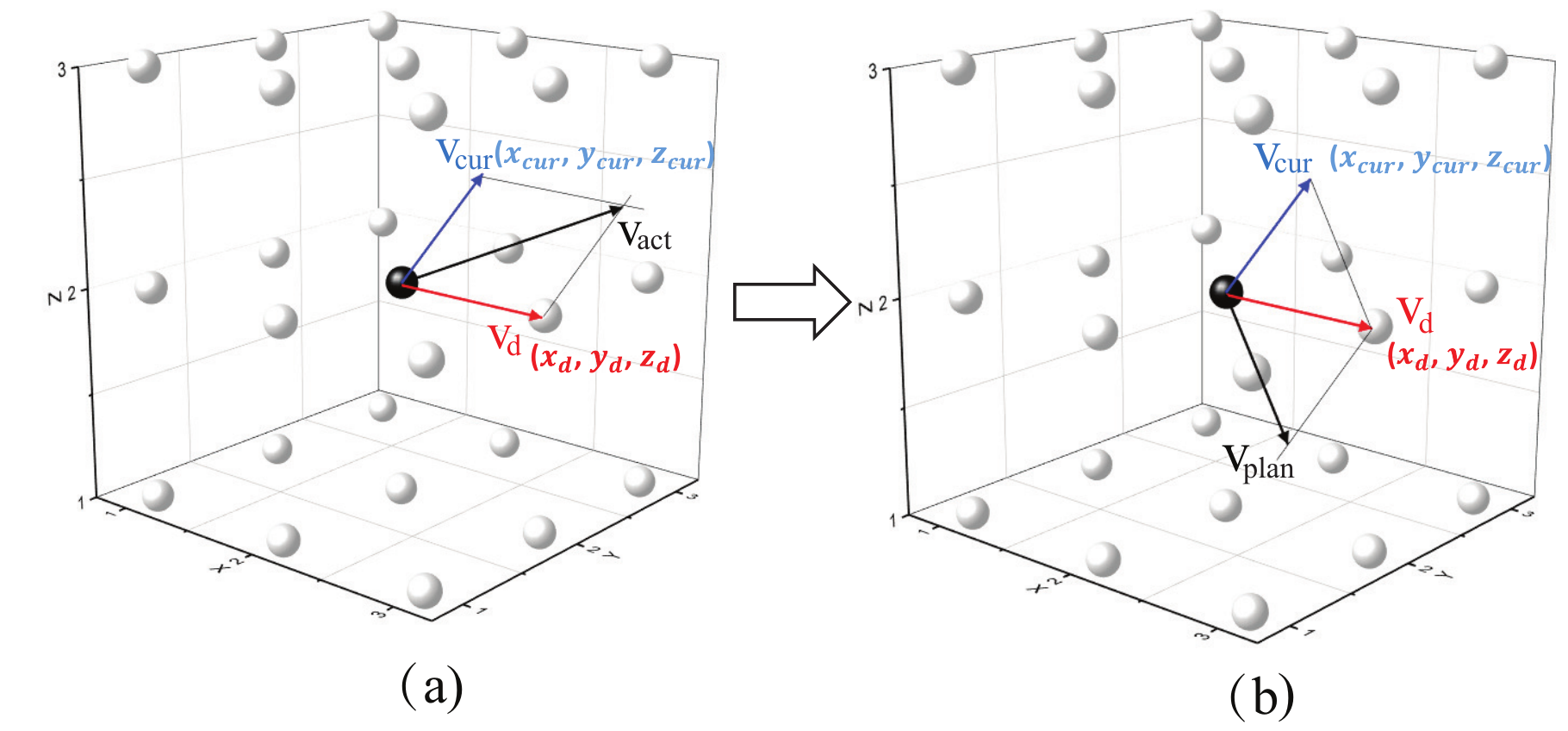}                         
        \caption*{Fig. 5. Elimination of the ocean current effect by introducing the third party vector $\mathbf{v_{plan}}$ in 3D modeling. (a) deviation induced without $\mathbf{v_{plan}}$; (b) no deviation with $\mathbf{v_{plan}}$.}	
\end{center} 
\end{figure}
\subsubsection{Current Effect-eliminated Bio-inspired Neural Network Path Planning (CBNNP) Algorithm}

The two separately designed components are combined together to obtain a final path planning algorithm that eliminates the current effect and sustains at the optimal path solution (see Fig. 6). The desired velocity and direction computed by the neural network algorithm at each position is passed to the elimination component to plan the velocity and direction the UUV should produce. The planned velocity and direction guarantee the vehicle to navigate along the optimal path by eliminating the current effect neuron by neuron, thus providing a robust and efficient path planning result for the UUV. The pseudocode for realizing the algorithm is given in Algorithm 1.
\begin{figure}[h]
\begin{center}
        \includegraphics[scale=0.38]{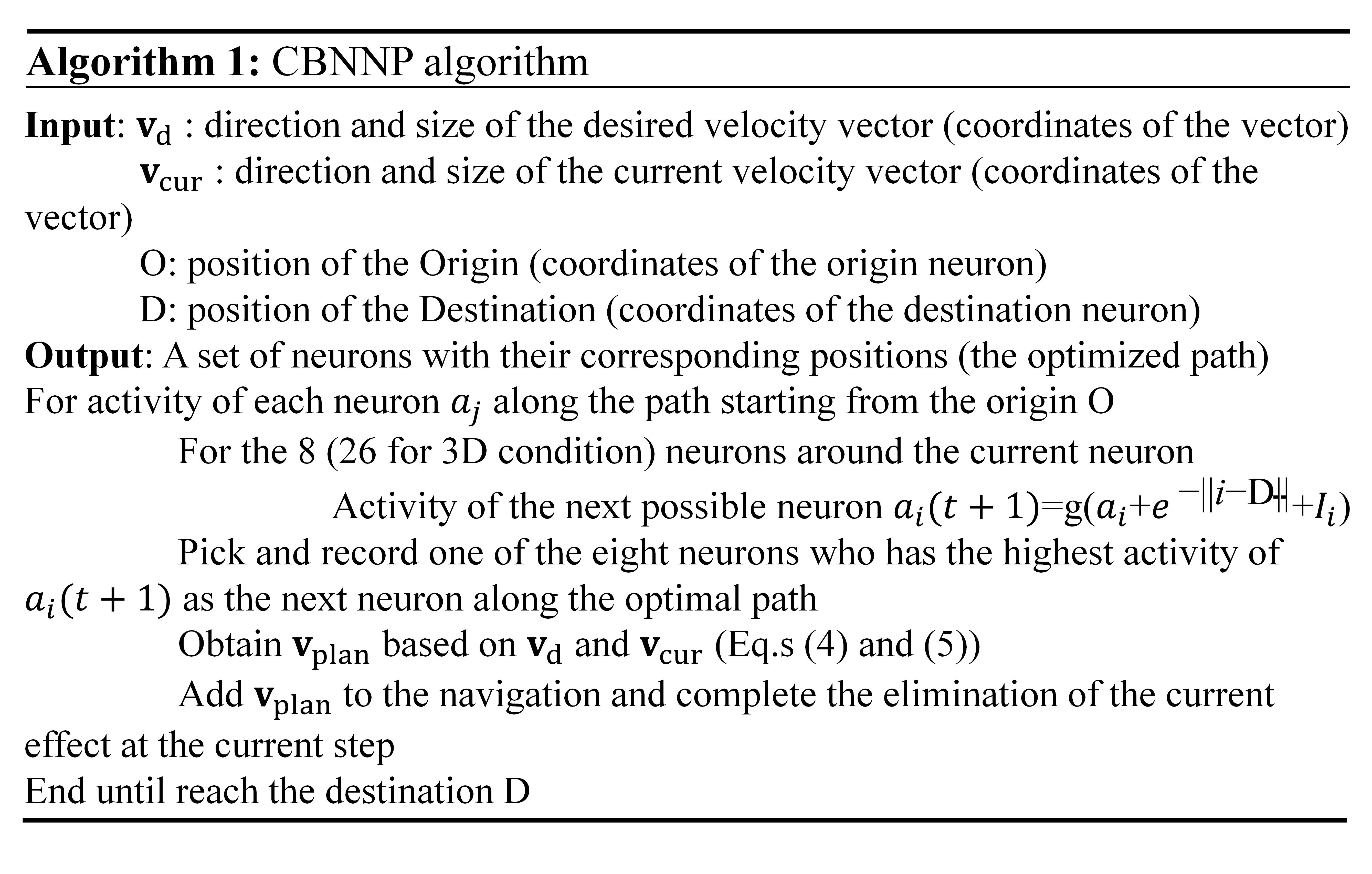}             
\end{center} 
\end{figure}

\begin{figure*}[h]
\begin{center}
        \includegraphics[scale=0.52]{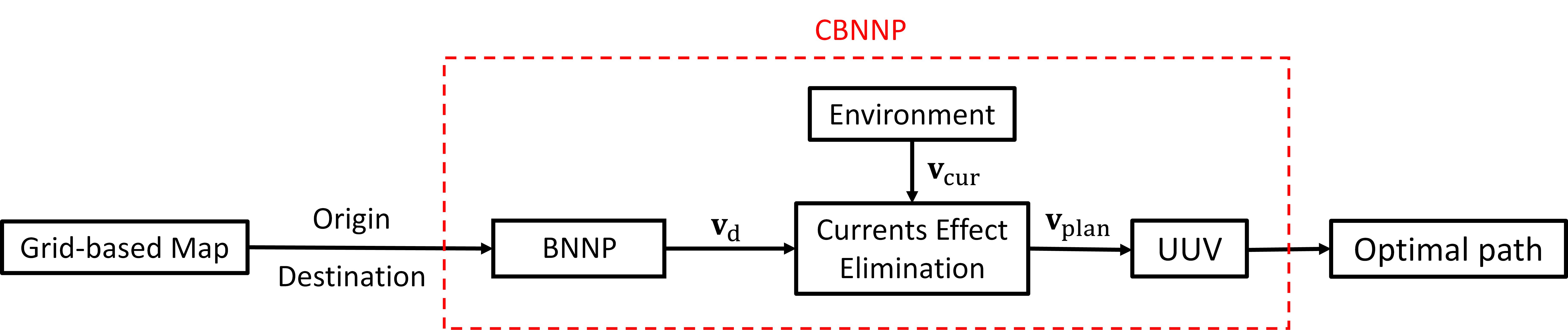}                         
        \caption*{Fig. 6. Schematic of the proposed optimal path planning method designed for the UUV in the underwater environment.}	
\end{center} 
\end{figure*}

\section{SIMULATION RESULTS AND ANALYSIS}
In this section, the path planning simulation results of the CBNNP are given under the conditions of different obstacle occupied ratios, currents of different directions and of different velocities. The simulation results of the BNNP algorithm that does not eliminate the effect of the ocean currents are also illustrated, and compared with the results controlled by the CBNNP algorithm. The distances of paths derived by both algorithms are listed and compared.
\subsection{2D Simulation Results}
The path planning simulation results in the two dimensional model of the CBNNP and the BNNP algorithm are given and analyzed in this section. The origin node lies at (2,1), while the destination node is at (9,9), which is to check the effectiveness of the CBNNP algorithm in a grid map of size $10\times10$. Initially, the desired velocity of the vehicle is consistently set at 1 m/s to work as a standard for all the simulation. The velocity of the static currents is set at 0.05 m/s. The constant coefficient $k_g$ is 0.5. The marks on the given figures all follow the same set of definitions.
\subsubsection{Results under Currents of Different Directions} 

\begin{figure}[ht]
\begin{center}
        \includegraphics[scale=0.42]{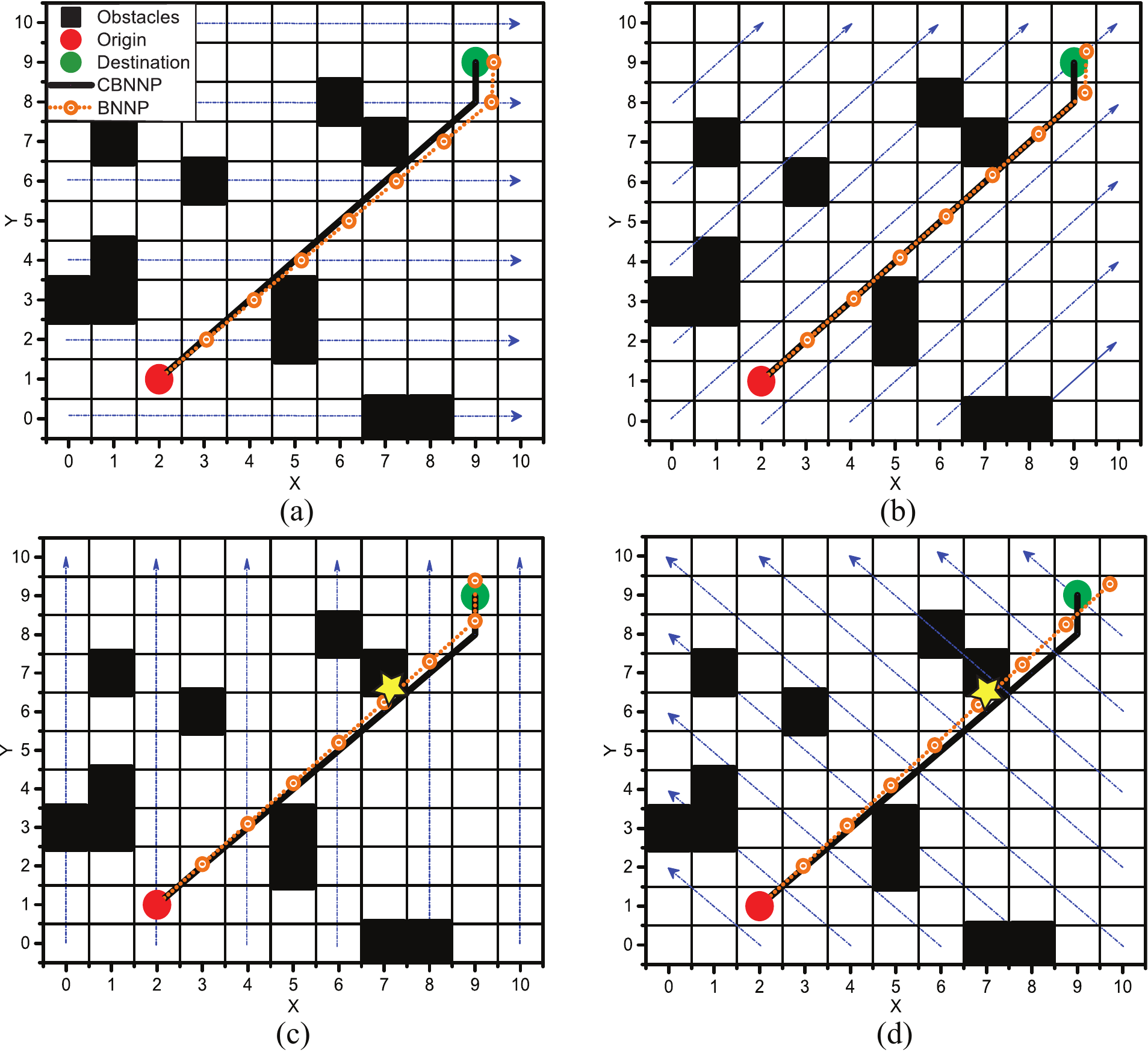}                         
        \caption*{Fig. 7. 2D comparison results of the paths given by the CBNNP algorithm and the BNNP algorithm under currents of different directions. Direction of the currents (all counterclockwise): (a) 0\degree; (b) 45\degree; (c) 90\degree; (d) 135\degree.}	
\end{center} 
\end{figure}

\begin{table*}[ht]
\centering
\captionsetup{justification=centering}
\caption*{TABLE \uppercase\expandafter{\romannumeral1}. 2D Moving Results of the Two Path Planning Algorithms under Currents of Different Directions \\(Unit: m, C: Collision, F: Fail to reach the destination.)}
\begin{tabular}{c||ccccc}

\hline
\diagbox  {Algorithms}{Direction of Currents (\degree)} & 0 & 45 & 90 & 135 & 180 \\
\hline
\hline
CBNNP & 10.8995 & 10.8995 & 10.8995 & 10.8995 & 10.8995\\
\hline
BNNP & 11.1512
& 11.2855 & C &  C \& F\ & C \& F\\
\hline
\end{tabular}
\end{table*}

In this section, currents of 0\degree, 45\degree, 90\degree and 135\degree counterclockwise are applied is to assess the effect of the fluid flow on the 2D map when tracking the path from the left bottom to the right bottom. The results of BNNP algorithm (in orange) have shown obvious deviations from the optimal planning path (see Fig. 7). Once the direction of the ocean currents changes, the deviation of the orange curve changes accordingly. For example, when the direction of the ocean currents is closer to the direction heading to the destination node, the deviation of the orange curve appears to be smaller due to the fact that the currents offers the vehicle more assistance to reach the destination (see Fig.s 7(a) and 7(b)). On the other hand, when the direction of the ocean currents gets farther from the vehicle’s supposed heading direction, the deviation becomes larger and follows the direction of the currents (Fig.s 7(c) and 7(d)). Under the cases where $\theta_{cur}$ is more than 90\degree from the heading direction, the vehicle deviates at the earlier stage, and cannot reach the destination within the size of this given map. At the same time, even under the existence of the ocean currents, the paths given by the CBNNP algorithm (in black line) all appear to have the shortest distance from the origin node to the destination node, and avoids all the possible obstacles on the map. Moreover, the CBNNP algorithm also successfully avoids the possibility of bumping into underwater obstacles (in yellow star) that may induce uncontrollable deviations if the effect of the ocean currents is not eliminated (Fig.s 7(c) and 7(d)). \\
\indent The distances of the paths controlled by the CBNNP and the BNNP algorithm are listed in Table \uppercase\expandafter{\romannumeral1}, with different directions of the currents applied. In general, the CBNNP algorithm performs a better result, where a robust sailing path is guaranteed due to the unchanged distances and all the obstacles are avoided. On the other hand, the BNNP algorithm fails to reach the supposed destination in most of the cases, where the vehicle either produces collisions or unrecoverable deviation (currents direction of 45\degree-180\degree). Moreover, under the cases when the vehicle successfully reaches the destination for both of the algorithms (currents direction of 0\degree and 45\degree), shorter distance is required by the CBNNP path. This supports the effectiveness of the CBNNP algorithm, which obviously saves the energy of the UUV while navigating in the underwater environment. 
\subsubsection{Results under Currents of Different Velocities}

\begin{figure}[ht]
\begin{center}
        \includegraphics[scale=0.42]{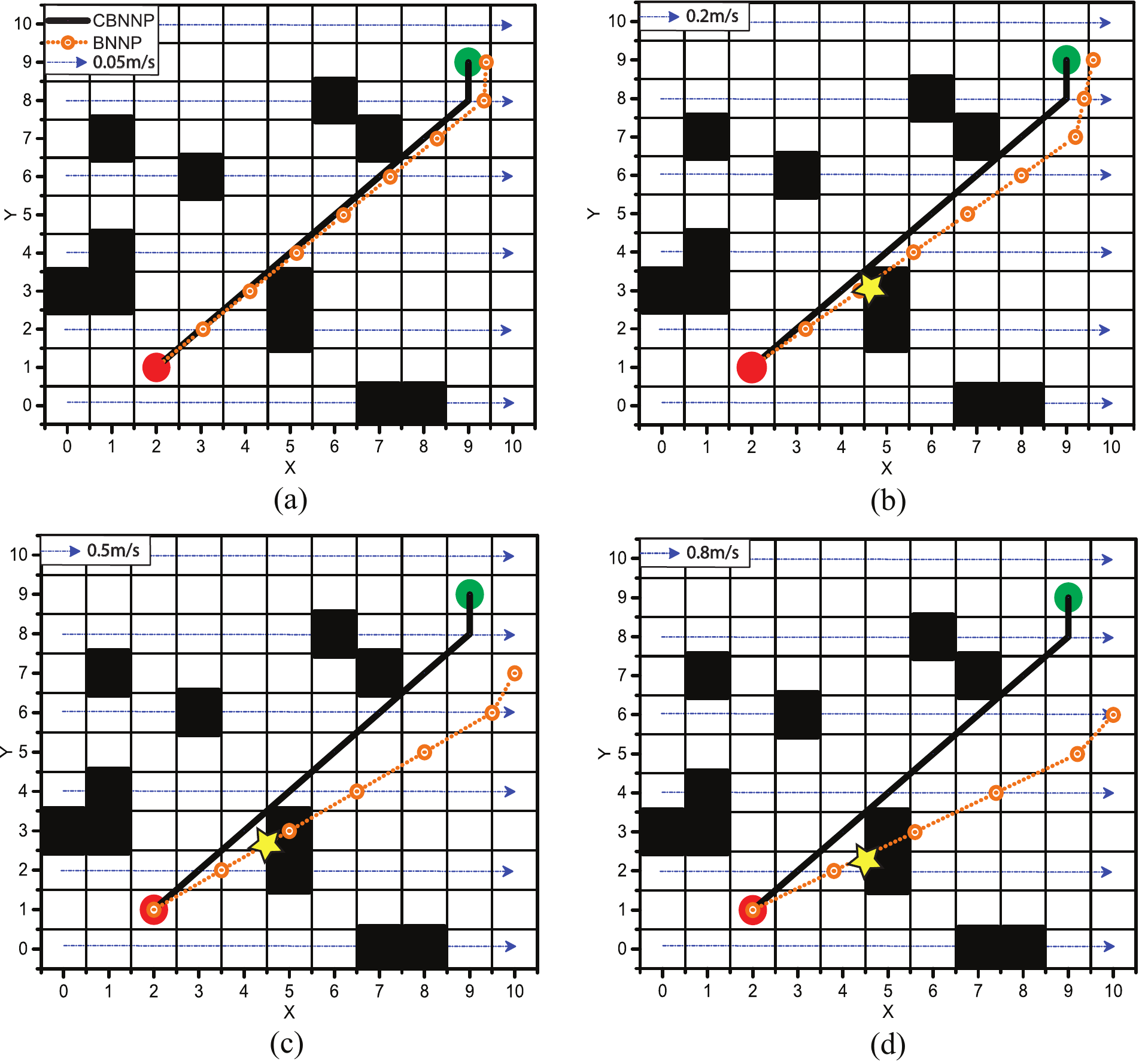}                         
        \caption*{Fig. 8. 2D comparison results of the paths given the CBNNP algorithm and the BNNP algorithm under currents of different velocities. Velocity of currents: (a) 0.05 m/s; (b) 0.2 m/s; (c) 0.5 m/s; (d) 0.8 m/s.}	
\end{center} 
\end{figure}

\begin{table*}[ht]
    \centering
    \captionsetup{justification=centering}
    \caption*{TABLE \uppercase\expandafter{\romannumeral2}. 2D Moving Results of the Two Path Planning Algorithms under Currents of Different Velocities \\(Unit: m, C: Collision, F: Fail to reach the destination.)}
    \setlength{\tabcolsep}{1mm}{
    \begin{tabular}{c||cccccccccc}
    \hline
    \diagbox  {Algorithms}{Velocity of Currents(m/s)} & 0.05 & 0.1 & 0.2 & 0.3 & 0.4 & 0.5 & 0.6 & 0.7 & 0.8 & 0.9 \\
    \hline
    \hline
    CBNNP & 10.8995 & 10.8995 & 10.8995 & 10.8995 & 10.8995 & 10.8995 & 10.8995 & 10.8995 & 10.8995 & 10.8995\\
    \hline
    BNNP & 11.1512 &
 F & C \& F & C \& F & C \& F & C \& F & C \& F & C \& F & C \& F & F \\
    \hline
    \end{tabular}}
\end{table*}

Considering the complex fluid status of the ocean currents when appearing underwater, the simulation results of the UUV path planning under the currents of different velocities are also illustrated. Same origin, destination position, $\mathbf{v_d}$ and $k_g$ are applied as the previous simulation, except the velocities of the currents are separately set at 0.05 m/s, 0.2 m/s, 0.5 m/s and 0.8 m/s and meanwhile the current direction is set at 0\degree counterclockwise for all cases to make a more direct comparison. It is obviously shown that the paths produced by the BNNP algorithm (in orange) deviate with the velocity of the currents (see Fig. 8). When the currents velocity increases, the deviation enlarges. The path given by the CBNNP algorithm successfully sustains at the most optimal solution, even under currents of high speed (Fig. 8(d)). Therefore, the CBNNP algorithm realizes the elimination of the current effect in the underwater environment with dynamic currents of different directions and velocities.\\
\indent According to the data of path distances illustrated in Table \uppercase\expandafter{\romannumeral2}, when the currents velocities increase, the path distance given by the CBNNP algorithm always sustains at the value of 10.8995, promising the robust optimal path planning result of the UUV. The paths under the control of the BNNP algorithm consume longer distance than the CBNNP algorithm (velocity of 0.005 m/s). This proves the inevitable effect of the ocean currents on the vehicle, where larger consumption of energy is needed to complete the path from the origin to destination. When the currents velocity becomes larger, the vehicle cannot accomplish the navigation to the destination, with large deviations and abrupt collisions (velocity of 0.2-0.8 m/s). The unpredictable failure of the BNNP method demonstrates the requirement of eliminating the ocean current effect, which is successfully achieved by the CBNNP algorithm.

\begin{figure}[ht]
\begin{center}
        \includegraphics[scale=0.42]{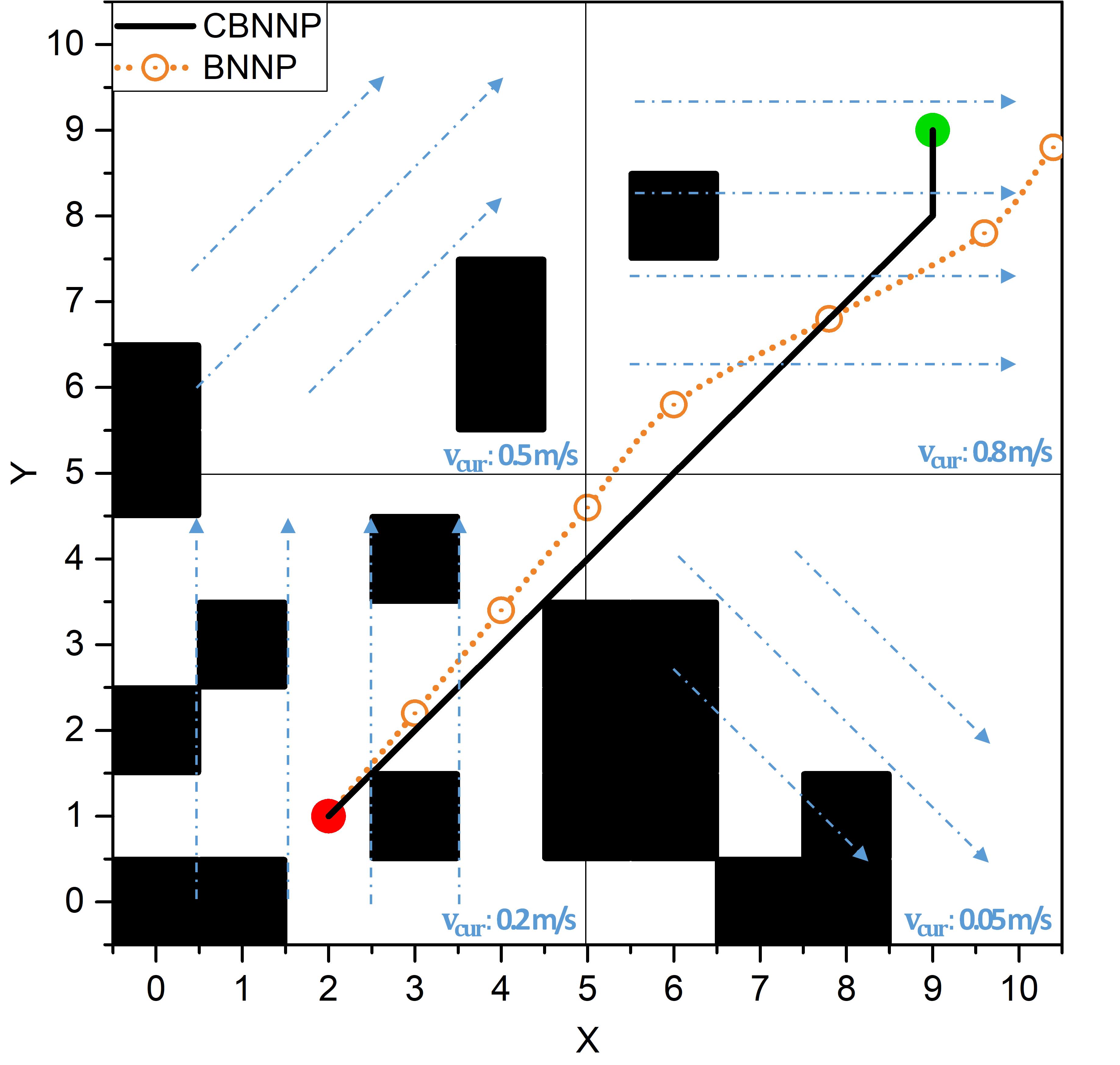}                         
        \caption*{Fig. 9. 2D comparison results of the paths given by the CBNNP algorithm and the BNNP algorithm under dynamic currents. $\mathbf{v}_{cur}$: velocity of currents.}	
\end{center} 
\end{figure}

The path planning results under the dynamic ocean currents with continuous changing velocities and directions are presented in Fig. 9. The path of the BNNP algorithm (in orange) vibrates fiercely with the currents flowing tendency, and finally out of control before reaching the destination (see the top-right part in Fig. 9). The CBNNP algorithm retains the most optimal path (in black line) throughout the whole process neglecting the dynamically changing currents. This proves the effectiveness of the CBNNP algorithm for eliminating the influence of the velocity of currents and successfully drives and tracks the most optimal path between the supposed origin and the destination.

\subsubsection{Results under Different Obstacle Occupied Ratios}
As the underwater obstacles are distributed randomly in both size and position, path planning simulation results of different obstacle occupied ratios are chosen to evaluate the effectiveness of the CBNNP algorithm. The obstacle occupied ratio refers to the approximate ratio between the size of all obstacles and the size of the map. Same origin, destination position, $\mathbf{v_d}$, $\mathbf{v_{cur}}$ and $k_g$ are applied as the previous simulation. The direction of the ocean currents are all set at 135\degree counterclockwise when the obstacle occupied ratio varies. 
\begin{figure}[ht]
\begin{center}
        \includegraphics[scale=0.42]{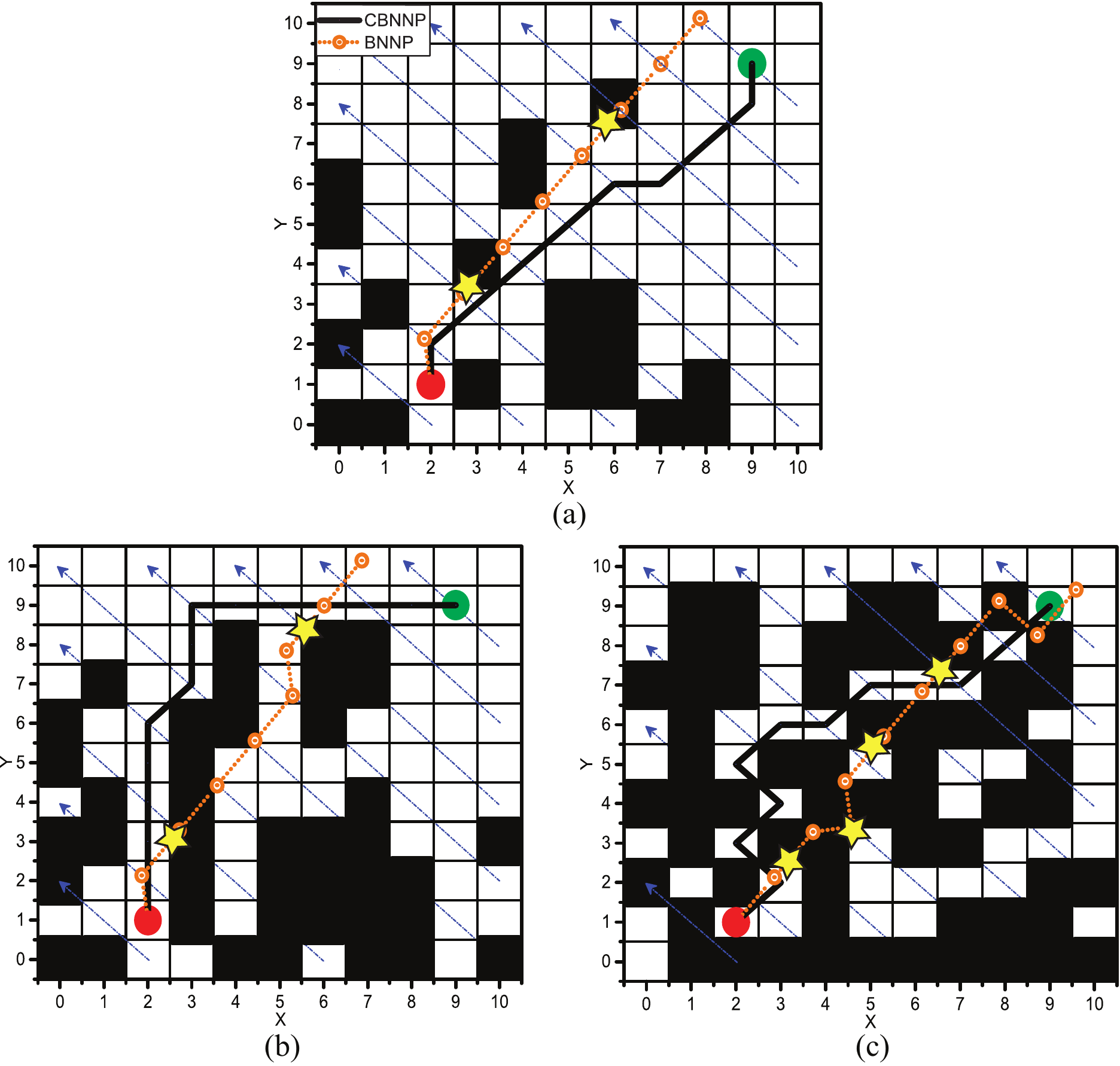}                         
        \caption*{Fig. 10. 2D comparison results of the paths given by the CBNNP algorithm and the BNNP algorithm under different obstacle occupied ratios: (a) 20$\%$ obstacles; (b) 40$\%$ obstacles; (c) 60$\%$ obstacles.}	
\end{center} 
\end{figure}\\

\begin{figure*}[t]
\begin{center}
        \includegraphics[scale=0.7]{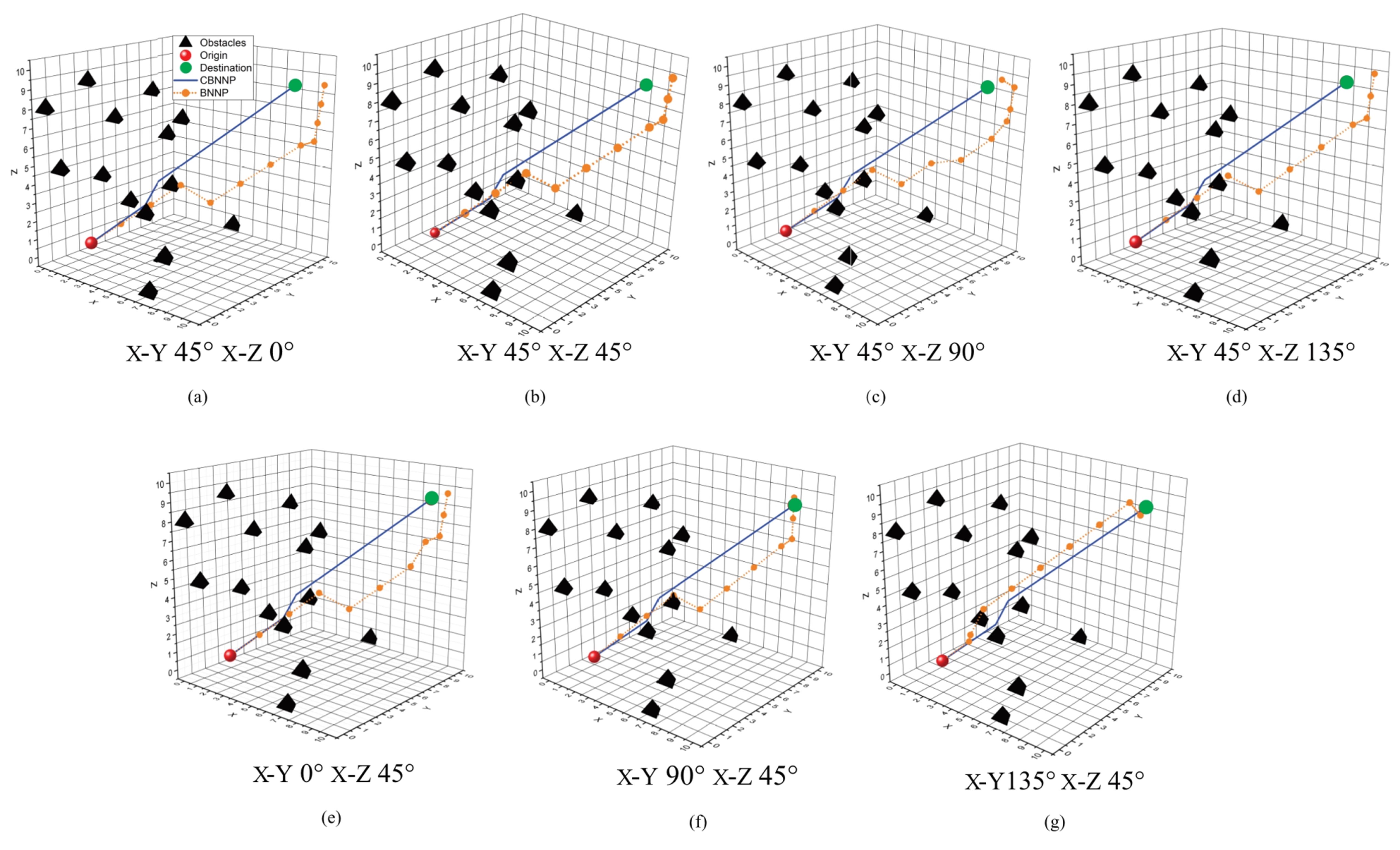}                         
        \caption*{Fig. 11. 3D comparison results of the paths given by the CBNNP and the BNNP algorithm under currents of different directions.}	
\end{center} 
\end{figure*}

\begin{table*}[t] 
    \centering
    \captionsetup{justification=centering}
    \caption*{TABLE \uppercase\expandafter{\romannumeral3}. 3D Moving Results of the Two Path Planning Algorithms under Currents of Different Directions\\(Unit: m, C: Collision, F: Fail to reach the destination.)}
    \begin{tabular}{c||ccccccccc}
    \hline
{Degree to X-Y plane} & 45 & 45 & 45 & 45 & 45 & 0 & 90 & 135 & 180 \\
\hline
{Degree to X-Z plane} & 0 & 45 & 90 & 135 & 180 & 45 & 45 & 45 & 45\\

    \hline
    \hline
    CBNNP & 13.5386 & 13.5386 & 13.5386 & 13.5386 & 13.5386 & 13.5386 & 13.5386 & 13.5386 & 13.5386\\
    \hline
    BNNP & C \& F & C \& F & C \& F & C \& F & C \& F & F & F & 16.3335 & F \\
    \hline
    \end{tabular}
\end{table*}

When the obstacle occupied ratio increases, the deviation of paths of the BNNP algorithm (in orange) performs the same increasing tendency (see Fig. 10). The optimal path is planned based on a continuous chain of excited neurons, where each neuron is chosen based on its previous neuron. Once the deviation is produced, the current neuron will create error messages that hold the wrong location information when passing it to the following neurons, thus leading to the accumulating deviation from the optimal path. Meanwhile the deviated paths perform more failures of avoiding the underwater obstacles (in yellow stars) as the obstacle occupied ratio increases. The CBNNP algorithm has the capability of correcting the location errors to keep the vehicle driving along the optimal path with the desired velocity and the desired heading direction, by deriving the corresponding velocity and direction that should be given by the vehicle. When the obstacle ratio increases, the paths controlled by the CBNNP algorithm show a satisfactory performance in avoiding all the obstacles, and the change of the obstacle distribution does not affect the accuracy of the CBNNP algorithm.

\subsection{3D Simulation Results}
 Considering the fact that the underwater environment is of three dimensions, the 3D simulation results of the CBNNP and the BNNP methods are demonstrated in this section. The origin is set at (2,1,1) while the destination lies at (9,9,9). Velocity value of $\mathbf{v_d}$ is assigned as 1 m/s and $\mathbf{v_{cur}}$ is 0.05 m/s. Coefficient $k_g$ is 0.5. The direction of the ocean currents is determined by two degrees, one is the degree to the X-Y plane, and the other is the degree to the X-Z plane. The size of the 3D map is $10\times10\times10$. The marks on the given figures all follow the same set of definitions.
\subsubsection{Results under Currents of Different Directions}

\begin{table*}[t] 
    \centering
    \captionsetup{justification=centering}
    \caption*{TABLE \uppercase\expandafter{\romannumeral4}. 3D Moving Results of the Two Path Planning Algorithms under Currents of Different Velocities\\
    (Unit: m, C: Collision, F: Fail to reach the destination.)}
    \setlength{\tabcolsep}{1mm}{
    \begin{tabular}{c||cccccccccc}
    \hline
    \diagbox  {Algorithms}{Velocity of Currents(m/s)}& 0.05 & 0.1 & 0.2 & 0.3 & 0.4 & 0.5 & 0.6 & 0.7 & 0.8 & 0.9 \\
    \hline
    \hline
    CBNNP & 13.5386 & 13.5386 & 13.5386 & 13.5386 & 13.5386 & 13.5386 & 13.5386 & 13.5386 & 13.5386 & 13.5386\\
    \hline
    BNNP &13.9122 & F & F & F & F & C \& F & F & F & F & F \\
    \hline
    \end{tabular}}
\end{table*}

The result of the CBNNP algorithm (in blue line) successfully completes the optimal path between the origin and the destination, and meanwhile avoids all the possible obstacles (in tetrahedron) distributed in the three-dimension space. The BNNP algorithm (in orange) produces deviations throughout the whole process, and finally fails to reach the destination in most cases. In the cases where the path controlled by the BNNP algorithm reaches the destination (see Fig. 11(g)), the distance it traveled is much longer than the path given by the CBNNP algorithm. The trajectory of the vehicle shown by the BNNP method also presents the deviation following the direction of the currents (see Fig. 11(g)). This illustrates that the existence of the ocean currents affects the optimization of the underwater path planning, bringing more unpredictability and energy consumption. Comparison results of the two algorithms drives the conclusion that the CBNNP is capable of accomplishing the optimal path planning by eliminating the current effect in the 3D environment, which supports the application of the CBNNP algorithm in the practical case of the UUV.\\
\indent The 3D path distance results are listed in Table \uppercase\expandafter{\romannumeral3}. Direction of currents in the 3D space is determined by its counterclockwise degree to the X-Y plane and to the X-Z plane. Compared to the results of the BNNP method, the paths given by the CBNNP algorithm all keep the vehicle navigate along the most optimal path, where the shortest distance is achieved. Under the control of the BNNP algorithm, the vehicle fails to reach the destination in most cases due to the inevitable deviation brought by the ocean currents. Additionally, longer distances are required for the BNNP method though it can reach the preset destination, which consumes more energy for the diving vehicle.

\subsubsection{Results under Currents of Different Velocities}


Comparison of the path planning results given by the CBNNP (in blue line) and the BNNP (in orange) algorithms under different currents velocities are presented in Fig. 12, where the currents direction is 0\degree to the X-Y plane and 45\degree to the X-Z plane. The vehicle controlled by the CBNNP algorithm sustains its trajectory along the most optimal path, neglecting the change of the velocity. It is obvious that the rate of the flow does not influence the optimization of the vehicle navigation, and all the possible obstacles are avoided under the CBNNP algorithm. However, for the BNNP algorithm, with the increment of the currents velocity, the paths it deducted never reach the destination and their deviation increases accordingly. The increasing deviation adds more unpredictability to the path planning, thus leading to the failure of the navigation and meanwhile consumes more time and energy to complete the operation.

The moving distance of the two algorithms under the currents of different velocities are listed in the Table \uppercase\expandafter{\romannumeral4}, where the direction of the currents are 0\degree to the X-Y plane and 45\degree to the X-Z plane. The same distance data given by the CBNNP algorithm for different currents velocities proves that the vehicle successfully accomplishes the most optimal path between the origin and the destination with the shortest length, regardless of the flow rate. While the BNNP algorithm fails to reach the destination in most cases, and the deviation rises due to the uncertainty brought by the increasing currents velocity. Even under the case controlled by the BNNP algorithm where the vehicle reaches the destination (see case of 0.05 m/s), the distance it traveled performs longer than the path given by the CBNNP algorithm, which supports the effectiveness of this refined algorithm.

\begin{figure}[ht]
\begin{center}
        \includegraphics[scale=0.4]{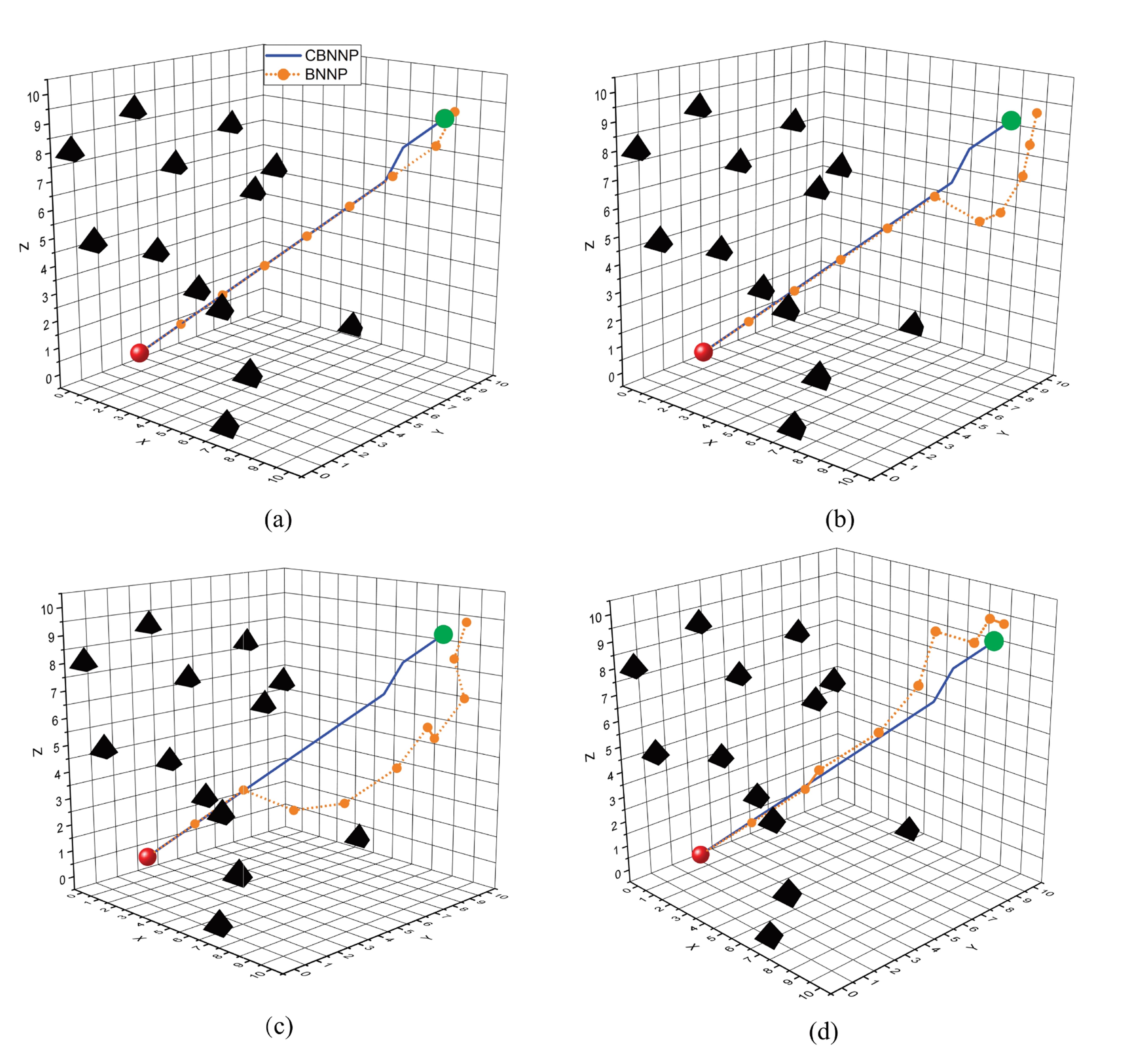}                         
        \caption*{Fig. 12. 3D comparison results of the paths given by the CBNNP and the BNNP algorithm under currents of different velocities: (a) 0.05 m/s; (b) 0.2 m/s; (c) 0.5 m/s; (d) 0.8 m/s.}	
\end{center} 
\end{figure}

\section{Conclusion}
In this paper, the problem of optimal path planning under the effect of ocean currents for the UUV is discussed. A bio-inspired neural network-based intelligent algorithm (CBNNP) is proposed to avoid collisions and eliminate the effect of currents. The path planning process is affected by currents of different directions and different velocities, where inevitable deviations are created and lead to the failure to follow the desired optimal path. Therefore, the CBNNP algorithm adds an adjusting component based on the parallelogram law, where it deducts the actual velocity and direction that the UUV should provide to balance off the deviation brought by the current effect step by step. The simulation results show that the CBNNP algorithm successfully eliminates the effect of ocean currents and avoids all possible collisions, neglecting the change of currents or the distribution of obstacles. This proves the effectiveness of the CBNNP algorithm in planning the optimal path for a UUV in the typical underwater environment, and meanwhile achieves lower energy consumption for the vehicle compared to the conventional method.


%




\ifCLASSOPTIONcaptionsoff
  \newpage
\fi



%

\scriptsize
\bibliographystyle{IEEEtran}
\bibliography{ref}

\begin{thebibliography}{10}
\providecommand{\url}[1]{#1}
\csname url@samestyle\endcsname
\providecommand{\newblock}{\relax}
\providecommand{\bibinfo}[2]{#2}
\providecommand{\BIBentrySTDinterwordspacing}{\spaceskip=0pt\relax}
\providecommand{\BIBentryALTinterwordstretchfactor}{4}
\providecommand{\BIBentryALTinterwordspacing}{\spaceskip=\fontdimen2\font plus
\BIBentryALTinterwordstretchfactor\fontdimen3\font minus
  \fontdimen4\font\relax}
\providecommand{\BIBforeignlanguage}[2]{{%
\expandafter\ifx\csname l@#1\endcsname\relax
\typeout{** WARNING: IEEEtran.bst: No hyphenation pattern has been}%
\typeout{** loaded for the language `#1'. Using the pattern for}%
\typeout{** the default language instead.}%
\else
\language=\csname l@#1\endcsname
\fi
#2}}
\providecommand{\BIBdecl}{\relax}
\BIBdecl

\bibitem{r1}
C.~{Hu}, R.~{Wang}, F.~{Yan}, and N.~{Chen}, ``Should the desired heading in
  path following of autonomous vehicles be the tangent direction of the desired
  path?'' \emph{IEEE Trans. Intell. Transp. Syst.}, vol.~16, no.~6, pp.
  3084--3094, 2015.

\bibitem{r2}
H.~L. Astri, T.~Elisabeth, B.~Peter, B.~Tommy, and D.~H. Kristian,
  ``\BIBforeignlanguage{English}{Managing risk in deepwater frontiers - key
  learnings from five continents},'' in \emph{\BIBforeignlanguage{English}{21st
  World Petroleum Congress, WPC 2014}}, Moscow, Russia, 2014.

\bibitem{r4}
S.~Gafurov and E.~Klochkov, ``\BIBforeignlanguage{English}{Autonomous unmanned
  underwater vehicles development tendencies},''
  \emph{\BIBforeignlanguage{English}{Procedia Eng.}}, vol. 106, pp. 141 -- 148,
  2015.

\bibitem{r5}
D.~Li, P.~Wang, and L.~Du, ``\BIBforeignlanguage{English}{Path planning
  technologies for autonomous underwater vehicles-a review},''
  \emph{\BIBforeignlanguage{English}{IEEE Access}}, vol.~7, pp. 9745 -- 9768,
  2019.

\bibitem{RN1346}
Y.-N. Ma, Y.-J. Gong, C.-F. Xiao, Y.~Gao, and J.~Zhang, ``Path planning for
  autonomous underwater vehicles: an ant colony algorithm incorporating alarm
  pheromone,'' \emph{IEEE Trans. Veh. Technol.}, vol.~68, no.~1, pp. 141--154,
  2019.

\bibitem{RN1338}
Y.~Singh, S.~Sharma, R.~Sutton, D.~Hatton, and A.~Khan, ``A constrained a*
  approach towards optimal path planning for an unmanned surface vehicle in a
  maritime environment containing dynamic obstacles and ocean currents,''
  \emph{Ocean Eng.}, vol. 169, pp. 187--201, 2018.

\bibitem{reviewuuvpp}
M.~Panda, B.~Das, B.~Subudhi, and B.~Pati, ``\BIBforeignlanguage{English}{A
  comprehensive review of path planning algorithms for autonomous underwater
  vehicles},'' \emph{\BIBforeignlanguage{English}{Int. J. Autom. Comput.}},
  vol.~17, no.~3, pp. 321 -- 352, Jun. 2020.

\bibitem{3dppauv}
M.~Aghababa, ``\BIBforeignlanguage{English}{3d path planning for underwater
  vehicles using five evolutionary optimization algorithms avoiding static and
  energetic obstacles},'' \emph{\BIBforeignlanguage{English}{Appl. Ocean
  Res.}}, vol.~38, pp. 48 -- 62, Oct. 2012.

\bibitem{pptime}
X.~Cao, C.~Sun, and M.~Chen, ``\BIBforeignlanguage{English}{Path planning for
  autonomous underwater vehicle in time-varying current},''
  \emph{\BIBforeignlanguage{English}{IEEE Trans. Intell. Transp. Syst.}},
  vol.~13, no.~8, pp. 1265 -- 1271, 2019.

\bibitem{3dpp}
B.~Li, R.~Zhao, G.~Xu, G.~Wang, Z.~Su, and Z.~Chen,
  ``\BIBforeignlanguage{English}{Three-dimensional path planning for an
  under-actuated autonomous underwater vehicle},'' in
  \emph{\BIBforeignlanguage{English}{Proceedings of the International Offshore
  and Polar Engineering Conference}}, vol.~1, Honolulu, HI, United states,
  2019, pp. 1518 -- 1524.

\bibitem{ppcon}
X.~Wang, X.~Yao, and L.~Zhang, ``\BIBforeignlanguage{English}{Path planning
  under constraints and path following control of autonomous underwater vehicle
  with dynamical uncertainties and wave disturbances},''
  \emph{\BIBforeignlanguage{English}{Journal of Intelligent and Robotic
  Systems: Theory and Applications}}, vol.~99, no. 3-4, pp. 891 -- 908, 2020.

\bibitem{Dijkstra}
E.~Dijkstra, ``\BIBforeignlanguage{English}{Communication with an automatic
  computer},'' Ph.D. dissertation, University of Amsterdam, Netherlands, 1959.

\bibitem{Astar1}
E.~H. Peter, J.~N. Nils, and R.~Bertram, ``\BIBforeignlanguage{English}{A
  formal basis for the heuristic determination of minimum cost paths},''
  \emph{\BIBforeignlanguage{English}{IEEE Trans. Syst. Sci. Cybern.}}, vol.
  SSC-4, no.~2, pp. 100--107, 1968.

\bibitem{uavcurrent2}
C.~S. Kulkarni and P.~F. Lermusiaux,
  ``\BIBforeignlanguage{English}{Three-dimensional time-optimal path planning
  in the ocean},'' \emph{\BIBforeignlanguage{English}{Ocean Modelling}}, vol.
  152, 2020.

\bibitem{wind1}
H.~Cao, H.~Cheng, and W.~Zhu, ``\BIBforeignlanguage{English}{Investigation of
  wind and sound field characteristics of multi-rotor unmanned aerial
  vehicle},'' \emph{\BIBforeignlanguage{English}{Noise and Vibration
  Worldwide}}, vol.~51, no. 7-9, pp. 158 -- 163, 2020.

\bibitem{wind2}
B.~H. Wang, D.~B. Wang, Z.~Ali, B.~T. Ting, and H.~Wang,
  ``\BIBforeignlanguage{English}{An overview of various kinds of wind effects
  on unmanned aerial vehicle},'' \emph{\BIBforeignlanguage{English}{Meas.
  Control}}, vol.~52, no. 7-8, pp. 731 -- 739, Sep. 2019.

\bibitem{wind4}
S.~Benders, A.~Wenz, and T.~Johansen, ``\BIBforeignlanguage{English}{Adaptive
  path planning for unmanned aircraft using in-flight wind velocity
  estimation},'' in \emph{\BIBforeignlanguage{English}{2018 International
  Conference on Unmanned Aircraft Systems (ICUAS). Proceedings}}, Piscataway,
  NJ, USA, 2018, pp. 483 -- 492.

\bibitem{RN1345}
Y.~Wu, K.~H. Low, and C.~Lv, ``Cooperative path planning for heterogeneous
  unmanned vehicles in a search-and-track mission aiming at an underwater
  target,'' \emph{IEEE Trans. Veh. Technol.}, vol.~69, no.~6, pp. 6782--6787,
  2020.

\bibitem{RN1337}
H.~Yu and T.~Su, ``A destination driven navigator with dynamic obstacle motion
  prediction,'' in \emph{IEEE International Conference on Robotics and
  Automation}, vol.~3, May 2001, Conference Proceedings, pp. 2692--2697.

\bibitem{rr3}
S.~Mahmoudzadeh, D.~Powers, and A.~Atyabi, ``\BIBforeignlanguage{English}{Uuv's
  hierarchical de-based motion planning in a semi dynamic underwater wireless
  sensor network},'' \emph{\BIBforeignlanguage{English}{IEEE Trans. Cybern.}},
  vol.~49, no.~8, pp. 2992 -- 3005, Aug. 2019.

\bibitem{RN1344}
M.~Hou, H.~Zhai, H.~Zhou, and F.~Zhang, ``Partitioning ocean flow field for
  underwater vehicle path planning,'' in \emph{OCEANS}.\hskip 1em plus 0.5em
  minus 0.4em\relax Marseille: IEEE, Jun. 2019.

\bibitem{EA1}
S.~MahmoudZadeh, D.~Powers, A.~Yazdani, K.~Sammut, and A.~Atyabi, ``Efficient
  auv path planning in time-variant underwater environment using differential
  evolution algorithm,'' \emph{J. Mar. Sci. Appl.}, vol.~17, no.~4, pp. 585 --
  591, Dec. 2018.

\bibitem{ant2}
G.~Han, Z.~Zhou, T.~Zhang, H.~Wang, L.~Liu, Y.~Peng, and M.~Guizani,
  ``Ant-colony-based complete-coverage path-planning algorithm for underwater
  gliders in ocean areas with thermoclines,'' \emph{IEEE Trans. Veh. Technol.},
  vol.~69, no.~8, pp. 8959 -- 8971, Aug. 2020.

\bibitem{EA2}
P.~Salgado and P.~Afonso, ``Evolutionary genes algorithm to path planning
  problems,'' in \emph{International Conference on Soft Computing and Pattern
  Recognition: Advances in Intelligent Systems and Computing}, Cham,
  Switzerland, 2020, pp. 217 -- 225.

\bibitem{bioins}
\BIBentryALTinterwordspacing
J.~Ni, L.~Wu, P.~Shi, and S.~X. Yang, ``\BIBforeignlanguage{English}{A dynamic
  bioinspired neural network based real-time path planning method for
  autonomous underwater vehicles},'' \emph{\BIBforeignlanguage{English}{Comput.
  Intell. Neurosci.}}, 2017. [Online]. Available:
  \url{https://doi.org/10.1155/2017/9269742}
\BIBentrySTDinterwordspacing

\bibitem{RN1348}
D.~Zhu, X.~Cao, B.~Sun, and C.~Luo, ``Biologically inspired self-organizing map
  applied to task assignment and path planning of an auv system,'' \emph{IEEE
  Trans. Cogn. Devel. Syst.}, vol.~10, no.~2, pp. 304--313, 2018.

\bibitem{Hoppp}
F.~Zanetti~de Castro and M.~Valle, ``A broad class of discrete-time
  hypercomplex-valued hopfield neural networks,'' \emph{Neural Netw.}, vol.
  122, pp. 54 -- 67, Feb. 2020.

\end{thebibliography}



%


\begin{IEEEbiography}
[{\includegraphics[width=1in,height=1.25in,clip,keepaspectratio]{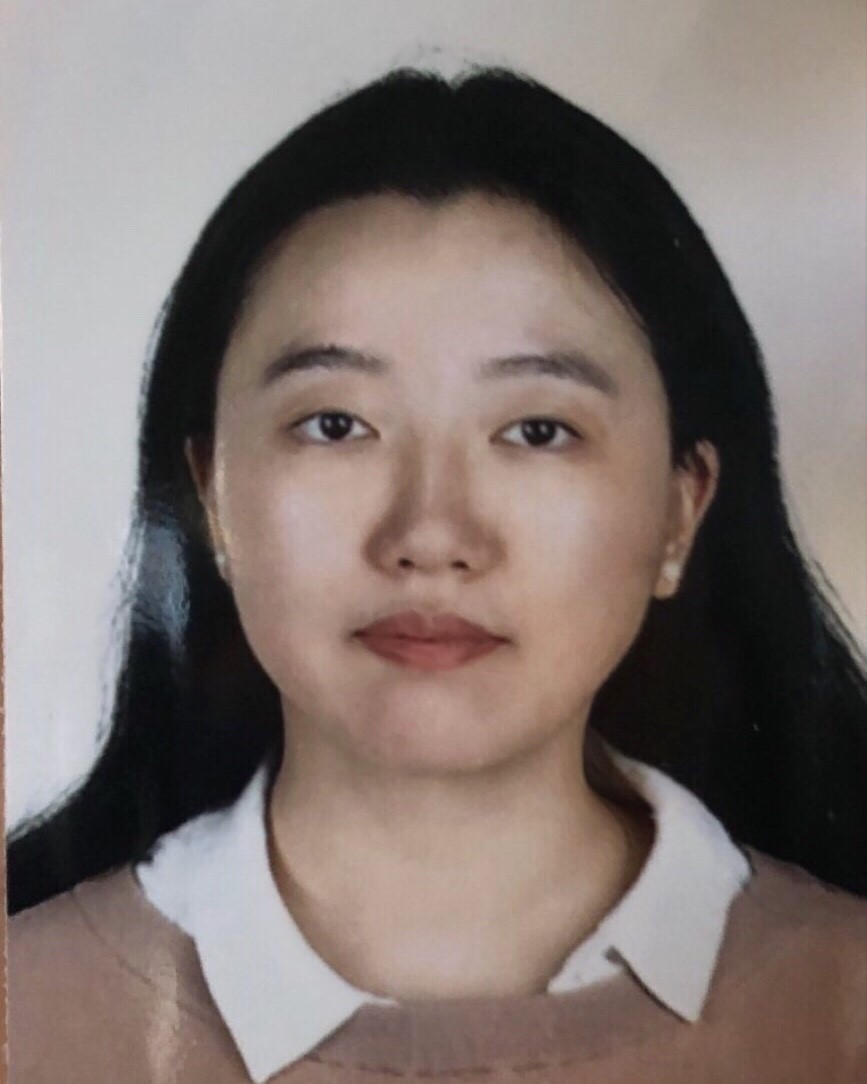}}]{Danjie Zhu} 
received the B.Sc. degree from Donghua University, Shanghai, China, in 2017; the M.Sc. degree from the University of California, Davis, U.S.A, in 2019. She is currently completing her Ph.D. degree at the University of Guelph, ON, Canada. Her research interest relates to the intelligent control of underwater robots.		
\end{IEEEbiography}

\begin{IEEEbiography}
[{\includegraphics[width=1in,height=1.25in,clip,keepaspectratio]{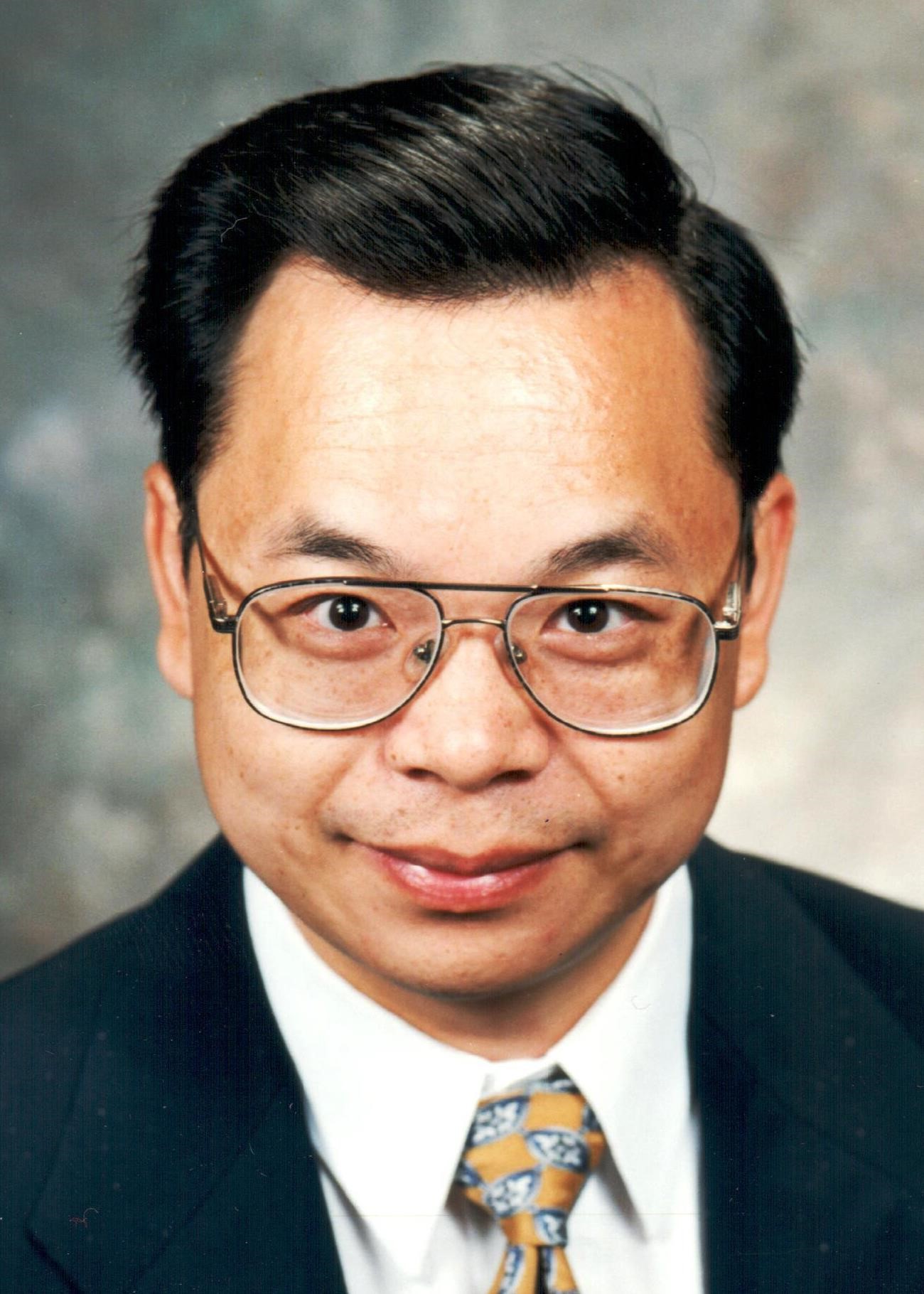}}]{Simon X. Yang} 
(S’97–M’99–SM’08) received the B.Sc. degree in engineering physics from Beijing University, Beijing, China, in 1987, the first of two M.Sc. degrees in biophysics from the Chinese Academy of Sciences, Beijing, in 1990, the second M.Sc. degree in electrical engineering from the University of Houston, Houston, TX, in 1996, and the Ph.D. degree in electrical and computer engineering from the University of Alberta, Edmonton, AB, Canada, in 1999.  Dr. Yang is currently a Professor and the Head of the Advanced Robotics and Intelligent Systems Laboratory at the University of Guelph, Guelph, ON, Canada. His research interests include robotics, intelligent systems, sensors and multi-sensor fusion, wireless sensor networks, control systems, machine learning, fuzzy systems, and computational neuroscience. 
Prof. Yang he has been very active in professional activities. He serves as the Editor-in-Chief of International Journal of Robotics and Automation, and an Associate Editor of IEEE Transactions on Cybernetics, IEEE Transactions of Artificial Intelligence, and several other journals. He has involved in the organization of many international conferences.
\end{IEEEbiography}








\end{document}